\pdfoutput=1

\documentclass[11pt]{article}

\usepackage[]{style}

\usepackage{times}
\usepackage{latexsym}

\usepackage[T1]{fontenc}

\usepackage[utf8]{inputenc}

\usepackage{microtype}

\usepackage{inconsolata}

\usepackage{graphicx}
\usepackage{amsmath}

\title{Discourse Structures Guided Fine-grained Propaganda Identification}

\author{Yuanyuan Lei and Ruihong Huang\\
        Department of Computer Science and Engineering\\
        Texas A\&M University, College Station, TX\\
        \texttt{\{yuanyuan, huangrh\}@tamu.edu}}

\begin{document}
\maketitle
\begin{abstract}

Propaganda is a form of deceptive narratives that instigate or mislead the public, usually with a political purpose. In this paper, we aim to identify propaganda in political news at two fine-grained levels: sentence-level and token-level. We observe that propaganda content is more likely to be embedded in sentences that attribute causality or assert contrast to nearby sentences, as well as seen in opinionated evaluation, speculation and discussions of future expectation. Hence, we propose to incorporate both local and global discourse structures for propaganda discovery and construct two teacher models for identifying PDTB-style discourse relations between nearby sentences and common discourse roles of sentences in a news article respectively. 
We further devise two methods to incorporate the two types of discourse structures for propaganda identification by either using teacher predicted probabilities as additional features or soliciting guidance in a knowledge distillation framework.  
Experiments on the benchmark dataset demonstrate that leveraging guidance from discourse structures can significantly improve both precision and recall of propaganda content identification.\footnote{The code and data link: https://github.com/yuanyuanlei-nlp/propaganda\_emnlp\_2023}

\end{abstract}

\section{Introduction}

\begin{figure*}[t]
  \centering
  \includegraphics[width = 6.3in]{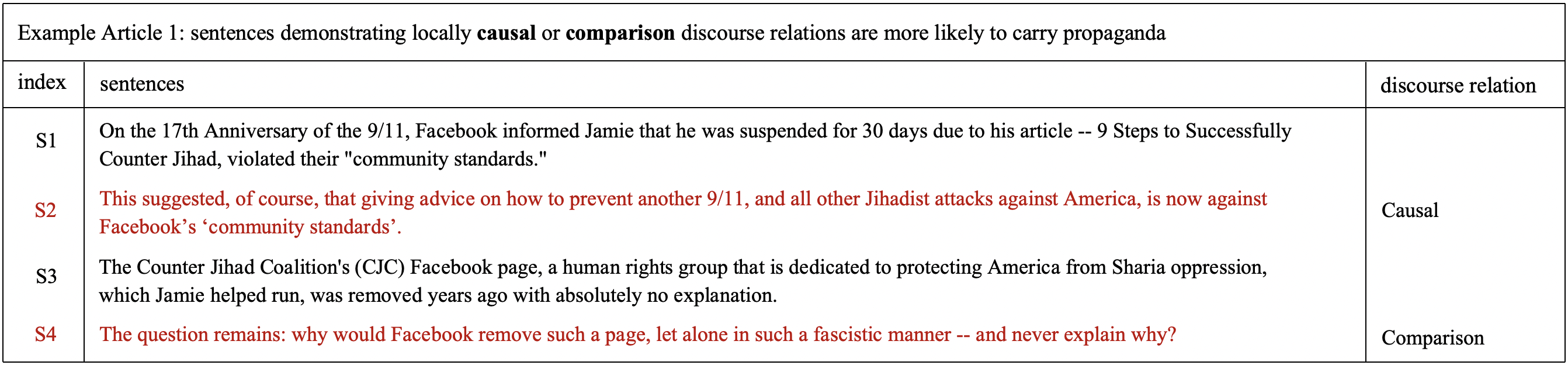}
  \caption{An example article containing propaganda sentences. Propaganda sentences are highlighted in \textcolor{red}{red}. S2 is a propaganda sentence showing deduction. S4 is a propaganda sentence proposing challenge or doubt. Their discourse relations with nearby sentence are shown in right column.}
  \label{introduction_example_discourse_relation}
\end{figure*}

\begin{figure*}[t]
  \centering
  \includegraphics[width = 6.3in]{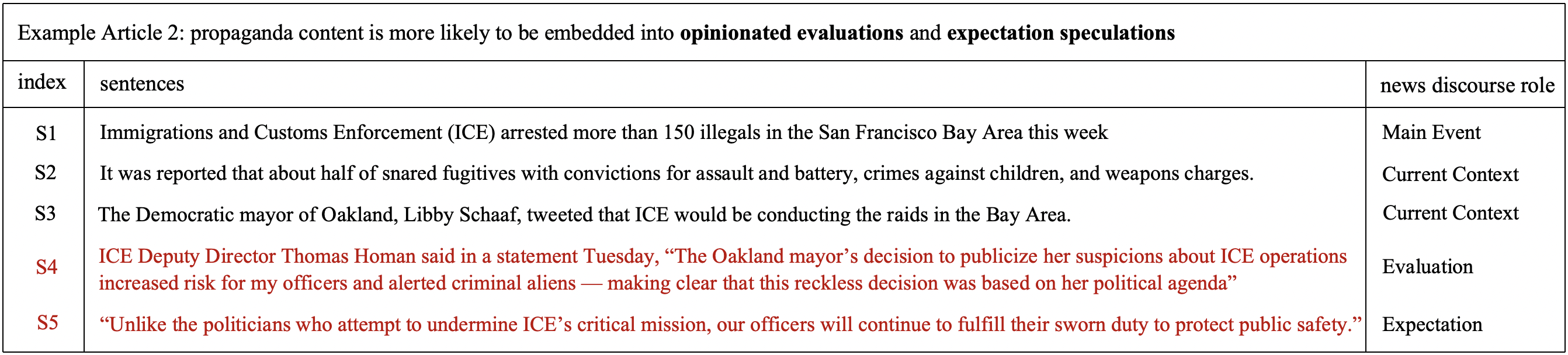}
  \caption{Another example article containing propaganda sentences, and the corresponding news discourse role of each sentence. Propaganda sentences are highlighted in \textcolor{red}{red}. Propaganda content is more likely to be embedded into opinionated evaluations and expectation speculations.}
  \label{introduction_example_discourse_role}
\end{figure*}

\begin{table*}[ht]
    \centering
    \scalebox{0.75}{
    \begin{tabular}{|c|cccc|c|}
        \hline
        & Comparison & Contingency & Temporal & Expansion & Total \\
        \hline
        propaganda & \textbf{102 (35.66)} & \textbf{146 (40.56)} & 18 (18.18) & 337 (32.13) & 620 (30.48) \\
        benign & 184 (64.34) & 214 (59.44) & 81 (81.82) & 712 (67.87) & 1414 (69.52) \\
        \hline
    \end{tabular}}
    \caption{The number (ratio) of propaganda and benign sentences that have each of the four discourse relations with nearby sentences. The ratio values higher than the overall ratio in the rightmost column are shown in \textbf{bold}.}
    \label{statistics_discourse_relation}
\end{table*}

Propaganda refers to a type of misleading and deceptive information used to promote or publicize a certain political point of view \cite{lasswell1927theory, henderson1943toward, stanley2015propaganda, rashkin2017truth}. This information is often manipulated in a strategic way to shape societal beliefs \cite{rashkin2017truth, barroncedeno2019proppy}. Propaganda can be harmful to both individuals and society as a whole, such as disseminating false information, inciting people's perceptions, leading to conflicts, perpetuating prejudices, impeding democracy process etc. \cite{bernays2005propaganda, stanley2015propaganda, little2017propaganda}. Despite its misleading nature and harmful effects, propaganda can be pervasive in political news, and has the potential to reach very large audiences \cite{glowacki2018news, tardaguila2018fake}. Hence, developing intelligent models to identify propaganda in political news is important and necessary.

Instead of detecting propaganda at the level of articles \cite{horne2018sampling, de-sarkar-etal-2018-attending, rashkin2017truth, rubin-etal-2016-fake}, this paper focuses on identifying propaganda at fine-grained levels: sentence-level and token-level. Through extracting the sentence or fragment imbued with propaganda, our aim is to accurately locate the propagandistic content and thereby provide detailed interpretable explanations. Propaganda content not only presents unverified or even false information, but also employs a variety of argumentation strategies to convince the readers \cite{yu-etal-2021-interpretable}. Therefore, identifying propaganda at fine-grained levels still remains a difficult and challenging task, and requires profound understanding of broader context in an article \cite{da-san-martino-etal-2019-fine}.

First, we observe that propaganda can be injected into sentences that attribute causality or assert contrast to nearby sentences. Take the article in Figure \ref{introduction_example_discourse_relation} as an example, the second sentence (S2) makes an illogical and misleading deduction from its preceding sentence: \textit{This suggested giving advice on how to prevent Jihadist attacks is now against community standards}. Propaganda content such as S2 usually disseminate the misinformation by leveraging causal relations, either by inferring baseless reasons or deducting speculative consequences. In addition, propaganda content can also utilize contrast relation to raise doubt or challenge credibility. For example, the last sentence (S4) casts doubts towards its previous sentence: \textit{why would Facebook remove such a page and never explain why?} Through the strategy of contrasting, the author aims to undermine the credibility of Facebook and thereby incite public protest. Accordingly, we propose that understanding the discourse relations of a sentence with its nearby sentences in the local context can enable discovery of propaganda contents.

Moreover, we observe that propaganda content or deceptive narratives is more likely to be embedded into opinionated evaluations or expectation speculations. In contrast, sentences describing factual occurrences are less likely to carry propaganda. Take the article in Figure \ref{introduction_example_discourse_role} as an example, the first three sentences, either reporting the main event or describing the current context triggered by the main event, all provide the readers with factual occurrences devoid of deceptive content. However, in the succeeding sentence (S4), the author includes a quotation to express emotional assessments: \textit{this reckless decision was based on her political agenda}. Propaganda sentences such as S4 always convince the readers and influence their mindset by inserting opinionated evaluations. Furthermore, the author speculates future expectations in the next sentence (S5) that \textit{ICE officers will continue to protect public safety}. Propaganda sentences such as S5 usually promise a bright yet unprovable future with the aim of gaining trust and support. Therefore, we propose that understanding the discourse role of a sentence in telling a news story can help reveal propaganda.

Motivated by the above observations, we propose to incorporate both local and global discourse structures for propaganda identification. Specifically, we establish two teacher models to recognize PDTB-style discourse relations between a sentence and its nearby sentences \cite{prasad-etal-2008-penn}, as well as identify one of eight common news discourse roles 
for each sentence based upon news discourse structure \cite{choubey-etal-2020-discourse}. We further devise two approaches to effectively incorporate the two types of discourse structures for propaganda identification. The first approach concatenates the predicted probabilities from two teacher models as additional features. The second approach develops a more sophisticated knowledge distillation framework, where we design a response-based distillation loss to mimic the prediction behavior of teacher models, as well as a feature relation-based distillation loss to seek guidance from the embeddings generated by teacher models. The response-based and feature relation-based distillation mutually complement each other, acquiring an enhanced guidance from discourse structures. Experiments on the benchmark dataset demonstrate the effectiveness of our approaches for leveraging discourse structures, with both precision and recall improved. The ablation study validates the necessity and synergy between local and global discourse structures.

\section{Discourse Structures}

In this section, we explain the details for the two discourse structures: discourse relation based on PDTB-style relations, and discourse role that draws upon news discourse structure. We also perform a statistical analysis to verify our empirical observations, and introduce the procedure of constructing teacher models for both discourse structures.

\subsection{Discourse Relations}

\subsubsection{PDTB Discourse Structure}

The Penn Discourse Treebank (PDTB) discourse structure \cite{prasad-etal-2008-penn} interprets the discourse relation between adjacent sentences in news articles into four types: 1). Comparison highlights prominent differences between two arguments, and represents the relation of contrasting or concession. 2). Contingency indicates two arguments causally influence each other, and represents a cause-and-effect or conditional relationship. 3). Temporal captures the temporal or chronological relationship between two arguments, such as precedence, succession, or simultaneously. 4). Expansion covers relations of elaborating additional details, providing explanations, or restating narratives.

\subsubsection{Statistical Analysis}

To validate the correlation between propaganda and discourse relations, we also conduct a statistical analysis on the validation set of propaganda dataset \cite{da-san-martino-etal-2019-fine}, where we run the model of classifying discourse relations. Table \ref{statistics_discourse_relation} shows the ratio of propaganda sentences that have each of the four discourse relations with nearby sentences. The numerical analysis confirms our observation: sentences that exhibit contingency and comparison relations with adjacent sentences are more prone to containing propaganda, whereas sentences that narrate events in a chronological order significantly contain less propaganda.

\subsubsection{Teacher Model for Discourse Relation}

We train the teacher model for discourse relations by using Longformer \cite{beltagy2020longformer} as the basic language model. The sentence pair embedding is the concatenation of hidden states at the two sentences start tokens <s>. A two-layer neural network is built on top of the pair embedding to predict discourse relations into comparison, contingency, temporal, or expansion. The model is trained on PDTB 2.0 data \cite{prasad-etal-2008-penn} that annotates both explicit and implicit relations between adjacent sentences. Considering propaganda sentences can be connected with nearby sentences with or without discourse connectives explicitly shown, we utilize both explicit and implicit discourse relations data for training.

Given a pair of sentences from the propaganda article, the local discourse relation teacher model generates the predicted probability of four relations between $i$-th sentence and its nearby sentence as:
\begin{equation}
    P_i^{local} = (P_{i1}^{local}, P_{i2}^{local}, P_{i3}^{local}, P_{i4}^{local})
\end{equation}

\begin{table*}[ht]
    \centering
    \scalebox{0.74}{
    \begin{tabular}{|c|cccccccc|c|}
        \hline
        & M1 & M2 & C1 & C2 & D1 & D2 & D3 & D4 & Total \\
        \hline
        propaganda & 66 (28.08) & 0 (none) & 9 (19.56) & 71 (17.07) & \textbf{51 (33.12)} & 3 (12.00) & \textbf{335 (42.84)} & \textbf{55 (36.18)} & 620 (30.48) \\
        benign & 169 (71.92) & 0 (none) & 37 (80.44) & 345 (82.93) & 103 (66.88) & 22 (88.00) & 447 (57.16) & 97 (63.82) & 1414 (69.52) \\
        \hline
    \end{tabular}}
    \caption{The number (ratio) of propaganda and benign sentences under each of the eight news discourse role types. The rightmost column shows the overall number (ratio). The ratio values higher than the overall ratio are shown in \textbf{bold}. M1: Main Event, M2: Consequence, C1: Previous Context, C2: Current Context, D1: Historical Event, D2: Anecdotal Event, D3: Evaluation, D4: Expectation}
    \label{statistics_discourse_role}
\end{table*}

\subsection{Discourse Role}

\subsubsection{News Discourse Structure}

The news discourse structure \cite{choubey-etal-2020-discourse} categorizes the discourse role of each sentence in news article into three broad types and eight subtypes: 1). main event contents contain two subtypes, Main event (M1) and Consequence (M2), and cover sentences that describe the main event and their immediate consequences which are often found inseparable from main events. 2). context-informing contents have two subtypes, Previous Event (C1) and Current Context (C2), and cover sentences that explain the context of the main event, including recent events and general circumstances. 3). additional supportive contents have four subtypes, describing past events that precede the main event in months and years (Historical Event (D1)) or unverifiable fictional situations (Anecdotal Event (D2)), or opinionated contents including reactions from immediate participants, experts, known personalities as well as journalists or news sources (Evaluation (D3)), except speculations and projected consequences referred as Expectation (D4).

\subsubsection{Statistical Analysis}

To verify the correlation between propaganda and news discourse structure, we perform a statistical analysis on the validation set of propaganda dataset \cite{da-san-martino-etal-2019-fine}, where we run the model of profiling news discourse structure \cite{choubey-huang-2021-profiling-news}. Table \ref{statistics_discourse_role} presents the ratio of propaganda sentences across the eight news discourse roles. The numerics validate our observations: propaganda is more likely to be embedded into sentences expressing opinions or evaluations (D3), speculating future expectations (D4), or fabricating historical background (D1). Conversely, sentences describing factual occurrences, such as reporting main event (M1) or informing context (C1, C2) are less likely to carry propaganda.

\subsubsection{Teacher Model for Discourse Role}

We follow the same framework in the current state-of-art model of profiling news discourse structure \cite{choubey-huang-2021-profiling-news}, where an actor-critic model is developed that selects between the standard REINFORCE \cite{williams1992simple} algorithm or imitation learning for training actor. Additionally, we replace the ELMo word embeddings \cite{peters-etal-2018-deep} with Longformer language model \cite{beltagy2020longformer}, which generates contextualized embeddings for long documents based on transformer \cite{vaswani2017attention} and provides further improvements to the current state-of-the-art.

Given a candidate propaganda article consisting of $n$ sentences, the global discourse role teacher model generates the predicted probability of eight discourse roles for $i$-th sentence as:
\begin{equation}
    P_i^{global} = (P_{i1}^{global}, P_{i2}^{global},...,P_{i8}^{global})
\end{equation}

\begin{figure*}[t]
  \centering
  \includegraphics[width = 6in, height = 3in]{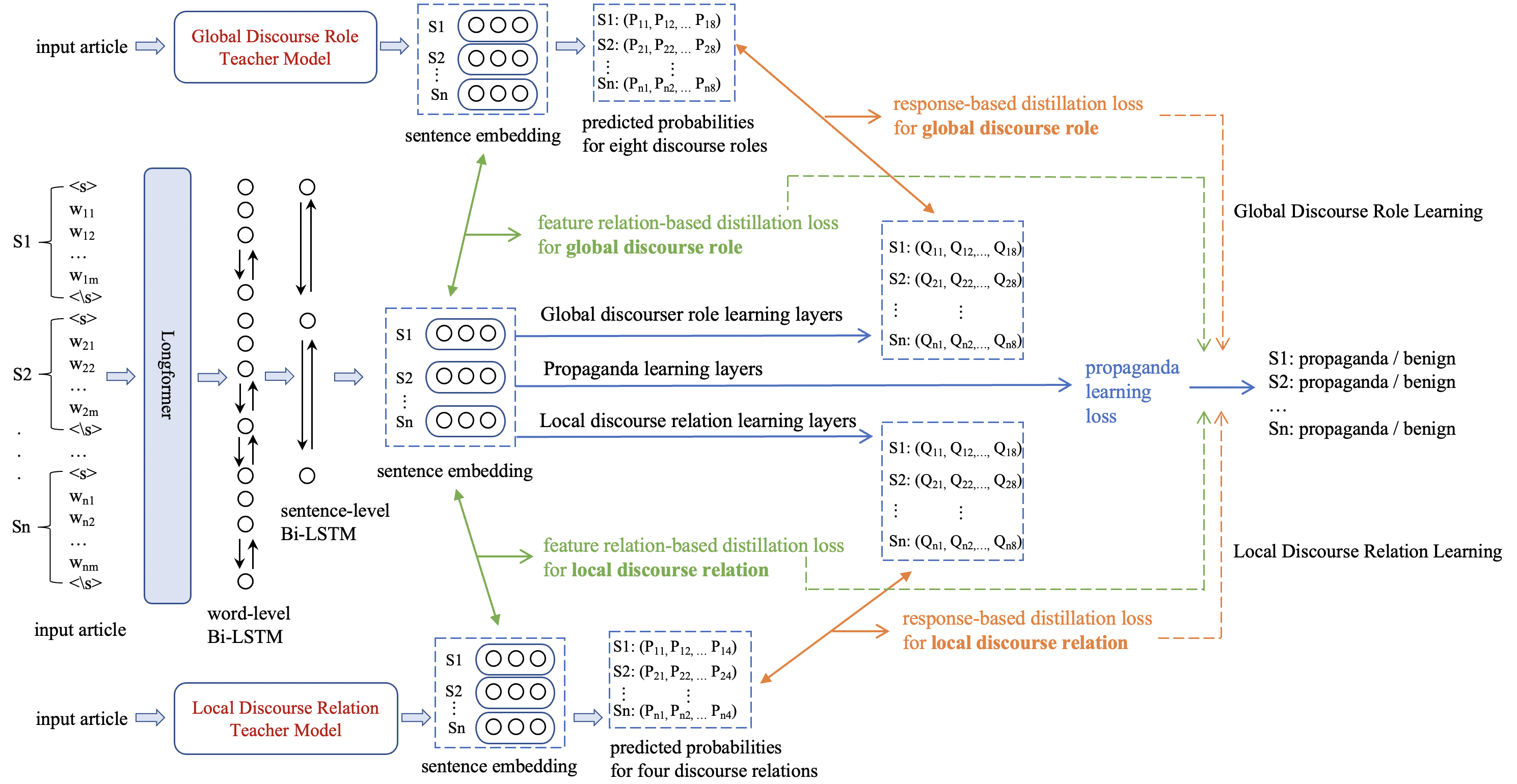}
  \caption{An illustration of propaganda identification guided by discourse structures via knowledge distillation}
  \label{methodology_figure}
\end{figure*}

\section{Fine-grained Propaganda Identification}

In order to incorporate the two types of discourse structures into propaganda identification, we further devise two methods: a feature concatenation model and a knowledge distillation model. Figure\ref{methodology_figure} illustrates the framework of knowledge distillation.

Considering the news articles are typically long, we utilized Longformer \cite{beltagy2020longformer} as the basic language model to encode the entire article. Given a candidate propaganda article consisting of $n$ sentences, sentence embeddings $(s_1, s_2,...,s_n)$ are initialized as the hidden state at sentence start tokens <s>. The $i$-th sentence contains $m$ tokens, and its tokens embeddings are $(w_{i1}, w_{i2},...w_{im})$.

\subsection{Feature Concatenation Model}

The feature concatenation model directly concatenates the predicted probabilities generated by the two teacher models as additional features, since they contain the discourse structures information. The updated feature vectors for $i$-th sentence and its $j$-th token in the two fine-grained tasks are:
\begin{equation}
    \small
    \begin{split}
        & \hat{s_i} = s_i \oplus P_i^{local} \oplus P_i^{global} \\
        & \hat{w_{ij}} = w_{ij} \oplus P_i^{local} \oplus P_i^{global}
    \end{split}
\end{equation}
where $\oplus$ denotes feature concatenation, $P_i^{local}$ and $P_i^{global}$ are probabilities of discourse relations and discourse roles predicted by two teacher models.

Additionally, a two-layer classification head is built on top of the updated embedding to make prediction. The cross-entropy loss is used for training.

\subsection{Knowledge Distillation Model}

The knowledge distillation model constructs additional learning layers to learn local discourse relation and global discourse role respectively. By optimizing the \textit{response-based distillation loss} to mimic the prediction behaviors of teacher, and the \textit{feature relation-based distillation loss} to learn from the embeddings generated by the teachers, the discourse structures information can be distilled into the task of propaganda identification.

\subsubsection{Learning Layers}

Three types of learning layers are built on top of sentence $s_i$ or token embedding $w_{ij}$: propaganda learning layer, student discourse relation learning layer, and student discourse role learning layer.

The \textit{propaganda learning layer} is to learn the main task of propaganda identification at either sentence level or token level:
\begin{equation}
    \small
    \begin{split}
        & Q_i^{propa} = softmax(W_2(W_1 s_i + b_1) + b_2) \\
        & Q_{ij}^{propa} = softmax(W_2(W_1 w_{ij} + b_1) + b_2)
    \end{split}
\end{equation}
where $Q_i^{propa}$ and $Q_{ij}^{propa}$ are the predicted probability of $i$-th sentence and its $j$-th token containing propaganda. $W_1, W_2, b_1, b_2$ are trainable parameters. The cross entropy loss is used for training:
\begin{equation}
    \small
    \begin{split}
        & Loss_{sent-propa} = -\sum_{i=1}^n P_i^{propa} \log(Q_i^{propa}) \\
        & Loss_{token-propa} = -\sum_{i=1}^n \sum_{j=1}^m P_{ij}^{propa} \log(Q_{ij}^{propa}) 
    \end{split}
\end{equation}
where $P_i^{propa}$ and $P_{ij}^{propa}$ are human annotated propaganda label for $i$-th sentence and its $j$-th token.

The \textit{student discourse relation learning layer} is built on top of the concatenation of $i$-th sentence embedding $s_i$ and its adjacent sentence embedding $s_{i-1}$, to learn the discourse relation between them from the teacher model:
\begin{equation}
    \small
    \begin{split}
        Q_i^{local} & = (Q_{i1}^{local}, Q_{i2}^{local},..., Q_{i4}^{local}) \\
        & = softmax(W_6(W_5 (s_i \oplus s_{i-1}) + b_5) + b_6)
    \end{split}
\end{equation}
where $W_5, W_6, b_5, b_6$ are trainable parameters in the student discourse relation layer, $Q_i^{local}$ is the learned outcome of predicting discourse relations.

The \textit{student discourse role learning layer} is built on top of the sentence embedding $s_i$, to learn its discourse role information from the teacher model:
\begin{equation}
    \small
    \begin{split}
        Q_i^{global} & = (Q_{i1}^{global}, Q_{i2}^{global},..., Q_{i8}^{global}) \\
        & = softmax(W_4(W_3 s_i + b_3) + b_4)
    \end{split}
\end{equation}
where $W_3, W_4, b_3, b_4$ are trainable parameters in the student discourse role layer, and $Q_i^{global}$ is its learned outcome of predicting eight discourse roles.

\subsubsection{Response-based Distillation}

The response-based distillation loss \cite{hinton2015distilling} is designed to minimize the discrepancy between the learned outcome of student layers and the predicted probability generated by the teacher models. By guiding the student layers to mimic the prediction behaviors of teachers, the knowledge of discourse relation and discourse role from the teachers can be distilled into the model.

Specifically, the Kullback–Leibler (KL) divergence loss is employed for measuring the distance between the learned probability of student layers and referenced probability from teacher models:
\begin{equation}
    \small
    \begin{split}
        Loss_{response-local} = \sum_{i=1}^n P_i^{local} \log\Big(\frac{P_i^{local}}{Q_i^{local}}\Big)
    \end{split}
\end{equation}
\begin{equation}
    \small
    \begin{split}
        Loss_{response-global} = \sum_{i=1}^n P_i^{global} \log\Big(\frac{P_i^{global}}{Q_i^{global}}\Big)
    \end{split}
\end{equation}
where $P_i^{local}$ and $P_i^{global}$ are response from the teachers, and are referenced as learning target. $Q_i^{local}$ and $Q_i^{global}$ are learned outcomes of student discourse relation layers and student discourse role layers. The response-based distillation loss penalizes the performance gap between teacher models and student layers, and forces student layers to be updated with discourse structures knowledge.

\begin{table*}[t]
    \centering
    \scalebox{0.9}{\begin{tabular}{|l|c|c|c|c||c|}
        \hline
        & Comparison & Contingency & Temporal & Expansion & Macro \\
        \hline
        Precision & 85.75 & 80.06 & 86.42 & 82.17 & 83.60 \\
        Recall & 83.73 & 74.93 & 90.32 & 85.38 & 83.59 \\
        F1-score & 84.73 & 77.41 & 88.33 & 83.75 & 83.55 \\
        \hline
    \end{tabular}}
    \caption{Performance of the PDTB discourse relations model (local discourse relation teacher) on PDTB 2.0 dataset.}
    \label{teacher_discourse_relation}
\end{table*}

\begin{table*}[t]
    \centering
    \scalebox{0.9}{\begin{tabular}{|l|c|c|c|c|c|c|c|c||c|}
        \hline
        & M1 & M2 & C1 & C2 & D1 & D2 & D3 & D4 & Macro \\
        \hline
        Precision & 55.56 & 37.88 & 43.72 & 67.21 & 66.67 & 62.69 & 75.22 & 62.15 & 63.23 \\
        Recall & 59.78 & 32.47 & 33.10 & 64.06 & 85.22 & 69.54 & 69.75 & 69.63 & 64.36 \\
        F1-score & 57.59 & 34.97 & 37.68 & 65.60 & 74.81 & 65.94 & 72.38 & 65.68 & 63.49 \\
        \hline
    \end{tabular}}
    \caption{Performance of the news discourse structure model (global discourse role teacher) on NewsDiscourse dataset. M1: Main Event, M2: Consequence, C1: Previous Context, C2: Current Context, D1: Historical Event, D2: Anecdotal Event, D3: Evaluation, D4: Expectation}
    \label{teacher_discourse_role}
\end{table*}

\subsubsection{Feature Relation-based Distillation}

The feature relation-based distillation loss is designed to seek guidance from the teacher-generated sentence embeddings which also contain discourse structures knowledge. However, sentence embedding itself has no absolute meaning and instead relies on its spatial relations with other contexts. Thus, rather than directly minimizing the euclidean distance between teacher-generated and student-learned features, we follow \cite{park2019relational} to guide the student layers to learn the spatial relations between sentences found in the teacher models.

Specifically, let $s_i^{local}$ and $s_i^{global}$ denotes the $i$-th sentence embedding trained by the two teachers. The spatial matrix of the teachers are computed:
\begin{equation}
    \small
    \begin{split}
        & M_{ik}^{local} = cosine(s_i^{local}, s_k^{local}) \\
        & M_{ik}^{global} = cosine(s_i^{global}, s_k^{global})
    \end{split}
\end{equation}
where $M_{ik}^{local}$ and $M_{ik}^{global}$ are spatial relation between $i$-th and $k$-th sentence in the teachers. Also, the spatial matrix of student-learned features is:
\begin{equation}
    \small
    \begin{split}
        & M_{ik} = cosine(s_i, s_k)
    \end{split}
\end{equation}
The feature relation-based distillation loss is the mean squared error (MSE) loss between spatial matrix of teacher models and student layers:
\begin{equation}
    \small
    \begin{split}
        & Loss_{relation-local} = \sum_{i,k} (M_{ik}^{local} - M_{ik})^2 \\
        & Loss_{relation-global} = \sum_{i,k} (M_{ik}^{global} - M_{ik})^2
    \end{split}
\end{equation}

To summarize, the response-based distillation and feature relation-based distillation mutually complement each other, with the former informed by teacher-predicted probabilities and the latter guided by teacher-generated embeddings.

\subsubsection{Learning Objective}

The total distillation loss for local discourse relation and global discourse role are:
\begin{equation}
    \small
    \begin{split}
        & Loss_{local} = Loss_{response-local} + Loss_{relation-local} \\
        & Loss_{global} = Loss_{response-global} + Loss_{relation-global}
    \end{split}
\end{equation}
The overall learning objective for identifying propaganda at sentence and token level are:
\begin{equation}
    \small
    \begin{split}
        & Loss_{sent} = Loss_{sent-propa} + Loss_{global} + Loss_{local} \\
        & Loss_{token} = Loss_{token-propa} + Loss_{global} + Loss_{local} \\
    \end{split}
\end{equation}

\begin{table*}[ht]
    \centering
    \scalebox{0.9}{
    \begin{tabular}{|l||ccc||ccc|}
        \hline
        & \multicolumn{3}{c||}{Sentence-level} & \multicolumn{3}{c|}{Token-level} \\
        \hline
        & Precision & Recall & F1 & Precision & Recall & F1 \\
        \hline
        Baseline Models & & & & & & \\
        all-propaganda & 24.86 & 100.00 & 39.82 & 10.41 & 100.00 & 18.86 \\
        chatgpt & 58.26 & 34.72 & 43.51 & 13.37 & 19.31 & 15.80 \\
        chatgpt + 5-shot & 56.42 & 37.34 & 44.94 & 14.68 & 20.84 & 17.22 \\
        chatgpt + discourse structures prompt & 57.93 & 38.61 & 46.34 & 15.82 & 21.96 & 18.39 \\
        \cite{da-san-martino-etal-2019-fine} & 63.20 & 53.16 & 57.74 & 39.57 & 36.42 & 37.90 \\
        \cite{da-san-martino-etal-2019-findings} & 60.28 & 66.48 & 63.23 & - & - & - \\
        \cite{fadel-etal-2019-pretrained} & - & - & 61.39 & - & - & - \\
        \cite{vlad-etal-2019-sentence} & 59.95 & 57.47 & 58.68 & - & - & - \\
        longformer & 60.32 & 60.50 & 60.41 & 34.60 & 39.81 & 37.03 \\
        \hline
        Feature Concatenation Models & & & & & & \\
        + local discourse relation & 61.72 & 62.09 & 61.90 & 35.38 & 41.92 & 38.37 \\
        + global discourse role & 61.50 & 63.58 & 62.52 & 36.39 & 41.28 & 38.68 \\
        + both discourse structures & 62.71 & 64.08 & 63.38 & 36.62 & 42.27 & 39.25 \\
        \hline
        Knowledge Distillation Models & & & & & & \\
        + local discourse relation & 60.40 & 66.17 & 63.15 & 35.18 & 43.49 & 38.90 \\
        + global discourse role & 61.88 & 66.86 & 64.27 & 37.65 & 43.32 & 40.28 \\
        + both discourse structures (full model) & \textbf{61.22} & \textbf{69.75} & \textbf{65.21} & \textbf{37.22} & \textbf{46.86} & \textbf{41.48} \\
        \hline
    \end{tabular}}
    \caption{Performance of sentence-level and token-level propaganda identification guided by discourse structures. Precision, Recall, and F1 of the propaganda class are shown. The model with the best performance is \textbf{bold}.}
    \label{propaganda_result}
\end{table*}

\section{Experiments}

\subsection{Dataset}

Acquiring human-annotated labels at fine-grained levels is challenging and expensive, leading to a limited resource of available datasets. In our subsequent experiments, we utilized the propaganda dataset published by \cite{da-san-martino-etal-2019-fine} that provides human-annotated labels for propaganda contents. We adhere to the same train / dev / test splitting in the released dataset. This propaganda dataset was also used in the NLP4IF-2019 challenge \cite{da-san-martino-etal-2019-findings}, which featured tasks involving sentence-level identification and token-level classification. In this paper, we specifically concentrate on propaganda identification at both the sentence and token levels.

\subsection{Teacher Models}

The teacher model for discourse relation is trained on PDTB 2.0 dataset \cite{prasad-etal-2008-penn}. Following its official suggestion, sections 2-21, sections 22 \& 24 and section 23 are used for train / dev / test set respectively. Table \ref{teacher_discourse_relation} displays the classification performance for the four discourse relations. On the other hand, the teacher model for discourse role is trained on News Discourse dataset \cite{choubey-etal-2020-discourse}. The performance of classifying the eight news discourse roles is presented in Table \ref{teacher_discourse_role}.

\subsection{Baseline Models}

We include the following baselines for comparison:
\begin{itemize}
    \item \textbf{all-propaganda}: a naive baseline that predicts all sentences / tokens into \textit{propaganda}
    \item \textbf{chatgpt}: an instruction prompt (\ref{prompt}) is designed for the large language model ChatGPT to automatically generate predicted labels for sentence / tokens in the same test set
    \item \textbf{chatgpt + 5-shot}: we add five examples of propaganda sentences and five examples of non-propaganda sentences into the prompt
    \item \textbf{chatgpt + discourse structures prompt}: we add the local discourse relation and global discourse role of each sentence into the prompt
    \item \cite{da-san-martino-etal-2019-findings}: we present the best performance achieved by the rank one team in the NLP4IF-2019 challenge, where the model was also trained on extensive corpora including Wikipedia and BookCorpus
    \item \cite{da-san-martino-etal-2019-fine}: where both sentence and token level propaganda identification tasks are performed
    \item \cite{fadel-etal-2019-pretrained}: pretrained ensemble learning is employed for sentence-level task
    \item \cite{vlad-etal-2019-sentence}: a capsule model architecture is designed for sentence-level task
    \item \textbf{longformer}: we build a baseline that follows the same framework and is equivalent to our developed model without discourse structures
\end{itemize}

\subsection{Experimental Setting}

The model takes the entire news article as input, and predicts the label for each sentence or token into \textit{propaganda} or \textit{benign}. The AdamW \cite{loshchilov2019decoupled} is used as the optimizer. The maximum length of input is set to 4096. The number of training epochs is 6. The learning rate is adjusted by a linear scheduler. The weight decay is set to be 1e-2. Precision, Recall, and F1 of \textit{propaganda} class is used as evaluation metric.

\begin{figure*}[ht]
  \centering
  \includegraphics[width = 6.3in]{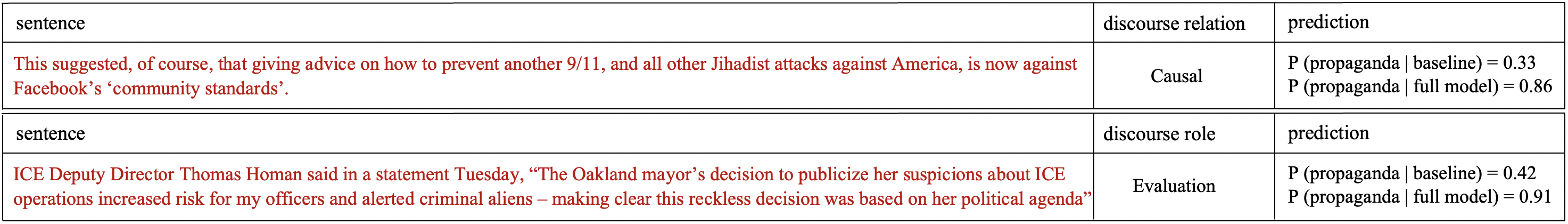}
  \caption{Examples of our method succeed in solving false negative error. Both the \textcolor{red}{red} sentences contain propaganda.}
  \label{analysis_example}
\end{figure*}

\subsection{Experimental Results}

Table \ref{propaganda_result} shows the performance of sentence-level and token-level propaganda identification.

Comparing feature concatenation models with the longformer baseline, we observe that integrating discourse relations or discourse roles as additional features brings consistent improvements for precision and recall, at both the sentence and token level tasks. This underscores that these two types of discourse structures can provide beneficial insights for identifying propaganda contents.

Comparing knowledge distillation models with the longformer baseline, it is evident that distilling the knowledge of discourse relations and discourse roles leads to a notable increase in recall by 9.25\% and a significant enhancement in F1 score by 4.8\%. Furthermore, in comparison to the previous best performance reported in \cite{da-san-martino-etal-2019-findings}, our knowledge distillation model exhibits superior performance in both precision and recall, ultimately achieving state-of-the-art results.

Comparing knowledge distillation models with feature concatenation models, we can see that distilling the knowledge from teacher models demonstrates stronger ability to incorporate two types of discourse structures, surpassing the approach of simply adding extra features.

Comparing our full model with the large language model ChatGPT, there still remains noticable performance gap, especially the recall. Also, the gap is even larger in terms of token-level task. Providing ChatGPT with extra examples or discourse structures information in the prompt can boost the performance a little bit, but it still remains inferior to our developed method.

\subsection{Ablation Study}

The ablation study of local discourse relation and global discourse role is also shown in Table \ref{propaganda_result}. Both the two types of discourse structures play an essential role in identifying propaganda content, at both the sentence and token level tasks. Incorporating the two discourse structures together can further boost recall and achieves the best performance.

\subsection{Effect of the Two Distillation Losses}

Moreover, we examine the effect of two types of distillation losses in Table \ref{two_distillation_losses}. Both response-based distillation and feature relation-based distillation yield substantial improvements. This demonstrates that learning from teacher-predicted probabilities and teacher-generated embeddings mutually complement each other, acquiring an enhanced guidance from discourse structures.

\begin{table}[ht]
    \centering
    \scalebox{0.9}{\begin{tabular}{|l|ccc|}
        \hline
        & Precision & Recall & F1 \\
        \hline
        longformer & 60.32 & 60.50 & 60.41 \\
        + response-based & 61.64 & 67.96 & 64.65 \\
        + relation-based & 60.40 & 66.77 & 63.42 \\
        + both (full model) & \textbf{61.22} & \textbf{69.75} & \textbf{65.21} \\
        \hline
    \end{tabular}}
    \caption{Ablation study of the two types of distillation losses: response-based and feature relation-based. Take sentence-level propaganda identification as an example.}
    \label{two_distillation_losses}
\end{table}

\subsection{Effect of the Four Local Discourse Relations}

In addition, we study the effect of the four local discourse relations in Table \ref{four_local_relations}. The results indicate that removing any one of the four discourse relations leads to a performance drop compared to the full model, as expected, the influence of expansion relations is relatively less compared to the other three types of relations.

\begin{table}[ht]
    \centering
    \scalebox{0.9}{\begin{tabular}{|l|ccc|}
        \hline
        & Precision & Recall & F1 \\
        \hline
        longformer & 60.32 & 60.50 & 60.41 \\
        the full model & \textbf{61.22} & \textbf{69.75} & \textbf{65.21} \\
        - comparison & 60.91 & 67.76 & 64.15 \\
        - contingency & 60.89 & 67.56 & 64.06 \\
        - temporal & 61.03 & 68.16 & 64.38 \\
        - expansion & 61.19 & 68.65 & 64.70 \\
        \hline
    \end{tabular}}
    \caption{Effect of removing each one of the four local discourse relations from the full model. Take sentence-level propaganda identification as an example.}
    \label{four_local_relations}
\end{table}

\subsection{Qualitative Analysis}

Figure \ref{analysis_example} presents examples of solving false negative error through the integration of discourse structures. The first propaganda sentence is inaccurately predicted as \textit{benign} by the longformer baseline. However, by incorporating the local causal discourse relation into the model, the prediction is corrected to \textit{propaganda}. Likewise, the second propaganda sentence is initially misclassified as a false negative by the baseline model. However, by leveraging the knowledge from the teacher model that this sentence plays an evaluation role in the article, the model successfully rectifies this error.

\section{Related Work}

\noindent\textbf{Propaganda} attracted research interests for years. Prior work focus on detecting propaganda at article-level \cite{rashkin2017truth, barroncedeno2019proppy}. The first work on fine-grained propaganda analysis was introduced by \cite{da-san-martino-etal-2019-fine, da-san-martino-etal-2019-findings}. A shared challenge focusing on token-level tasks was launched by \cite{da-san-martino-etal-2020-semeval}. Several approaches have been developed for propaganda analysis, such as \cite{vlad-etal-2019-sentence} designed an unified neural network, \cite{fadel-etal-2019-pretrained} utilized pretrained ensemble learning, \cite{dimitrov-etal-2021-detecting} trained a multimodal model mixing textual and visual features, and \cite{vijayaraghavan-vosoughi-2022-tweetspin} employed multi-view representations. In this paper, we focus on identifying propaganda in news articles at both sentence-level and token-level, leveraging discourse structures.

\noindent\textbf{Misinformation Detection} was also studied for years, such as fake news \cite{perez-rosas-etal-2018-automatic, oshikawa-etal-2020-survey}, rumor \cite{wei-etal-2021-towards, li-etal-2019-rumor}, political bias \cite{baly-etal-2020-detect, chen-etal-2020-analyzing}, and logical fallacy \cite{jin-etal-2022-logical, alhindi-etal-2022-multitask}. Although propaganda may intersect with fake news, political bias, and logical fallacies, however, they are all distinct phenomena and tasks. Fake news and rumor always hallucinate untruthful information. Political bias refers to selectively reporting verified facts while leaving readers to arrive at their own conclusions. Logical fallacies focus on errors in reasoning and argumentations to reach an invalid conclusion. In contrast, propaganda presents unverified speculation or projections in the same tone as facts, and employs a variety of persuasion strategies to convince the readers, with the purpose to manipulate public beliefs to a predetermined conclusion.

\noindent\textbf{Media Bias}. In the most broad sense, propaganda news articles is a type of biased news reports. However, media bias often refers to ideological bias these days \cite{kiesel2019semeval, fan2019plain, lei-huang-2022-shot}, and ideological bias is often expressed in a subtle way or under a neutral tone \cite{van2020context, lei-etal-2022-sentence} by selectively including certain facts to subtly shift public opinions \cite{fan2019plain}. In contrast, propaganda is not limited to hyper-partisan cases and can be applied to influence public beliefs in a way that aligns with the interests of the propagandist \cite{stanley2015propaganda, rashkin2017truth}. Propaganda often contains intensely emotional or opinionated content to incite or persuade the public \cite{da-san-martino-etal-2019-fine}, or presents unverified speculations, projections and deceptions \cite{miller2019propaganda, brennen2017making}. Indeed, in the current media landscape, ideologically biased media sources and propaganda media sources are often labeled separately, for example, Media Bias/Fact Check\footnote{\url{https://mediabiasfactcheck.com/}} distinguishes ideologically biased sources, conspiracy theory sources, questionable sources which includes major propaganda sources, and a couple other categories. Ideology bias and propaganda are studied separately as well in the NLP community \cite{barroncedeno2019proppy, liu-etal-2022-politics}, and each task features their own benchmark datasets \cite{fan2019plain,baly-etal-2020-detect,da-san-martino-etal-2019-fine} with documents retrieved from different media sources.

\section{Conclusion}

This paper aims to identify propaganda at sentence-level and token-level. We propose to incorporate two types of discourse structures into propaganda identification: local discourse relation and global discourse role. We further design a feature concatenation model and a knowledge distillation model to leverage the guidance from discourse structures.

\section*{Limitations}

This paper specifically concentrates on the identification of propaganda as a specific form of misinformation. There still exists various other forms of misinformation, such as fake news, conspiracy theories, and more. While the designed discourse structures method has demonstrated its usefulness in identifying propaganda, its effectiveness for other types of misinformation remains unknown. 

\section*{Ethics Statement}

This paper focuses on the detection of propaganda, which falls within the broader category of misinformation and disinformation. 
The release of code and models should be utilized for the purpose of combating misinformation and not for spreading further misinformation.

\section*{Acknowledgments}

We would like to thank the anonymous reviewers for their valuable feedback and input. We gratefully acknowledge support from National Science Foundation via the awards IIS-1942918 and IIS-2127746.


\bibliography{anthology,custom}

\begin{thebibliography}{46}
\expandafter\ifx\csname natexlab\endcsname\relax\def\natexlab#1{#1}\fi

\bibitem[{Alhindi et~al.(2022)Alhindi, Chakrabarty, Musi, and
  Muresan}]{alhindi-etal-2022-multitask}
Tariq Alhindi, Tuhin Chakrabarty, Elena Musi, and Smaranda Muresan. 2022.
\newblock \href {https://aclanthology.org/2022.emnlp-main.560} {Multitask
  instruction-based prompting for fallacy recognition}.
\newblock In \emph{Proceedings of the 2022 Conference on Empirical Methods in
  Natural Language Processing}, pages 8172--8187, Abu Dhabi, United Arab
  Emirates. Association for Computational Linguistics.

\bibitem[{Baly et~al.(2020)Baly, Da~San~Martino, Glass, and
  Nakov}]{baly-etal-2020-detect}
Ramy Baly, Giovanni Da~San~Martino, James Glass, and Preslav Nakov. 2020.
\newblock \href {https://doi.org/10.18653/v1/2020.emnlp-main.404} {We can
  detect your bias: Predicting the political ideology of news articles}.
\newblock In \emph{Proceedings of the 2020 Conference on Empirical Methods in
  Natural Language Processing (EMNLP)}, pages 4982--4991, Online. Association
  for Computational Linguistics.

\bibitem[{Barron-Cedeno et~al.(2019)Barron-Cedeno, Da~San~Martino, Jaradat, and
  Nakov}]{barroncedeno2019proppy}
Alberto Barron-Cedeno, Giovanni Da~San~Martino, Israa Jaradat, and Preslav
  Nakov. 2019.
\newblock Proppy: A system to unmask propaganda in online news.
\newblock In \emph{Proceedings of the 33rd AAAI Conference on Artificial
  Intelligence}, pages 9847--9848, Honolulu, HI, USA. AAAI Press.

\bibitem[{Beltagy et~al.(2020)Beltagy, Peters, and
  Cohan}]{beltagy2020longformer}
Iz~Beltagy, Matthew~E. Peters, and Arman Cohan. 2020.
\newblock \href {http://arxiv.org/abs/2004.05150} {Longformer: The
  long-document transformer}.

\bibitem[{Bernays(2005)}]{bernays2005propaganda}
Edward~L Bernays. 2005.
\newblock \emph{Propaganda}.
\newblock Ig publishing.

\bibitem[{Brennen(2017)}]{brennen2017making}
Bonnie Brennen. 2017.
\newblock Making sense of lies, deceptive propaganda, and fake news.
\newblock \emph{Journal of Media Ethics}, 32(3):179--181.

\bibitem[{Chen et~al.(2020)Chen, Al~Khatib, Wachsmuth, and
  Stein}]{chen-etal-2020-analyzing}
Wei-Fan Chen, Khalid Al~Khatib, Henning Wachsmuth, and Benno Stein. 2020.
\newblock \href {https://doi.org/10.18653/v1/2020.nlpcss-1.16} {Analyzing
  political bias and unfairness in news articles at different levels of
  granularity}.
\newblock In \emph{Proceedings of the Fourth Workshop on Natural Language
  Processing and Computational Social Science}, pages 149--154, Online.
  Association for Computational Linguistics.

\bibitem[{Choubey and Huang(2021)}]{choubey-huang-2021-profiling-news}
Prafulla~Kumar Choubey and Ruihong Huang. 2021.
\newblock \href {https://doi.org/10.18653/v1/2021.findings-emnlp.137}
  {{P}rofiling news discourse structure using explicit subtopic structures
  guided critics}.
\newblock In \emph{Findings of the Association for Computational Linguistics:
  EMNLP 2021}, pages 1594--1605, Punta Cana, Dominican Republic. Association
  for Computational Linguistics.

\bibitem[{Choubey et~al.(2020)Choubey, Lee, Huang, and
  Wang}]{choubey-etal-2020-discourse}
Prafulla~Kumar Choubey, Aaron Lee, Ruihong Huang, and Lu~Wang. 2020.
\newblock \href {https://doi.org/10.18653/v1/2020.acl-main.478} {Discourse as a
  function of event: Profiling discourse structure in news articles around the
  main event}.
\newblock In \emph{Proceedings of the 58th Annual Meeting of the Association
  for Computational Linguistics}, pages 5374--5386, Online. Association for
  Computational Linguistics.

\bibitem[{Da~San~Martino et~al.(2019{\natexlab{a}})Da~San~Martino,
  Barr{\'o}n-Cede{\~n}o, and Nakov}]{da-san-martino-etal-2019-findings}
Giovanni Da~San~Martino, Alberto Barr{\'o}n-Cede{\~n}o, and Preslav Nakov.
  2019{\natexlab{a}}.
\newblock \href {https://doi.org/10.18653/v1/D19-5024} {Findings of the
  {NLP}4{IF}-2019 shared task on fine-grained propaganda detection}.
\newblock In \emph{Proceedings of the Second Workshop on Natural Language
  Processing for Internet Freedom: Censorship, Disinformation, and Propaganda},
  pages 162--170, Hong Kong, China. Association for Computational Linguistics.

\bibitem[{Da~San~Martino et~al.(2020)Da~San~Martino, Barr{\'o}n-Cede{\~n}o,
  Wachsmuth, Petrov, and Nakov}]{da-san-martino-etal-2020-semeval}
Giovanni Da~San~Martino, Alberto Barr{\'o}n-Cede{\~n}o, Henning Wachsmuth,
  Rostislav Petrov, and Preslav Nakov. 2020.
\newblock \href {https://doi.org/10.18653/v1/2020.semeval-1.186}
  {{S}em{E}val-2020 task 11: Detection of propaganda techniques in news
  articles}.
\newblock In \emph{Proceedings of the Fourteenth Workshop on Semantic
  Evaluation}, pages 1377--1414, Barcelona (online). International Committee
  for Computational Linguistics.

\bibitem[{Da~San~Martino et~al.(2019{\natexlab{b}})Da~San~Martino, Yu,
  Barr{\'o}n-Cede{\~n}o, Petrov, and Nakov}]{da-san-martino-etal-2019-fine}
Giovanni Da~San~Martino, Seunghak Yu, Alberto Barr{\'o}n-Cede{\~n}o, Rostislav
  Petrov, and Preslav Nakov. 2019{\natexlab{b}}.
\newblock \href {https://doi.org/10.18653/v1/D19-1565} {Fine-grained analysis
  of propaganda in news article}.
\newblock In \emph{Proceedings of the 2019 Conference on Empirical Methods in
  Natural Language Processing and the 9th International Joint Conference on
  Natural Language Processing (EMNLP-IJCNLP)}, pages 5636--5646, Hong Kong,
  China. Association for Computational Linguistics.

\bibitem[{De~Sarkar et~al.(2018)De~Sarkar, Yang, and
  Mukherjee}]{de-sarkar-etal-2018-attending}
Sohan De~Sarkar, Fan Yang, and Arjun Mukherjee. 2018.
\newblock \href {https://aclanthology.org/C18-1285} {Attending sentences to
  detect satirical fake news}.
\newblock In \emph{Proceedings of the 27th International Conference on
  Computational Linguistics}, pages 3371--3380, Santa Fe, New Mexico, USA.
  Association for Computational Linguistics.

\bibitem[{Dimitrov et~al.(2021)Dimitrov, Bin~Ali, Shaar, Alam, Silvestri,
  Firooz, Nakov, and Da~San~Martino}]{dimitrov-etal-2021-detecting}
Dimitar Dimitrov, Bishr Bin~Ali, Shaden Shaar, Firoj Alam, Fabrizio Silvestri,
  Hamed Firooz, Preslav Nakov, and Giovanni Da~San~Martino. 2021.
\newblock \href {https://doi.org/10.18653/v1/2021.acl-long.516} {Detecting
  propaganda techniques in memes}.
\newblock In \emph{Proceedings of the 59th Annual Meeting of the Association
  for Computational Linguistics and the 11th International Joint Conference on
  Natural Language Processing (Volume 1: Long Papers)}, pages 6603--6617,
  Online. Association for Computational Linguistics.

\bibitem[{Fadel et~al.(2019)Fadel, Tuffaha, and
  Al-Ayyoub}]{fadel-etal-2019-pretrained}
Ali Fadel, Ibraheem Tuffaha, and Mahmoud Al-Ayyoub. 2019.
\newblock \href {https://doi.org/10.18653/v1/D19-5020} {Pretrained ensemble
  learning for fine-grained propaganda detection}.
\newblock In \emph{Proceedings of the Second Workshop on Natural Language
  Processing for Internet Freedom: Censorship, Disinformation, and Propaganda},
  pages 139--142, Hong Kong, China. Association for Computational Linguistics.

\bibitem[{Fan et~al.(2019)Fan, White, Sharma, Su, Choubey, Huang, and
  Wang}]{fan2019plain}
Lisa Fan, Marshall White, Eva Sharma, Ruisi Su, Prafulla~Kumar Choubey, Ruihong
  Huang, and Lu~Wang. 2019.
\newblock In plain sight: Media bias through the lens of factual reporting.
\newblock \emph{arXiv preprint arXiv:1909.02670}.

\bibitem[{Glowacki et~al.(2018)Glowacki, Narayanan, Maynard, Hirsch, Kollanyi,
  Neudert, Howard, Lederer, and Barash}]{glowacki2018news}
Monika Glowacki, Vidya Narayanan, Sam Maynard, Gustavo Hirsch, Bence Kollanyi,
  Lisa-Maria Neudert, Phil Howard, Thomas Lederer, and Vlad Barash. 2018.
\newblock News and political information consumption in mexico: Mapping the
  2018 mexican presidential election on twitter and facebook.
\newblock Technical Report COMPROP DATA MEMO 2018.2, Oxford University, Oxford,
  UK.

\bibitem[{Henderson(1943)}]{henderson1943toward}
Edgar~H Henderson. 1943.
\newblock Toward a definition of propaganda.
\newblock \emph{The Journal of Social Psychology}, 18(1):71--87.

\bibitem[{Hinton et~al.(2015)Hinton, Vinyals, and Dean}]{hinton2015distilling}
Geoffrey Hinton, Oriol Vinyals, and Jeff Dean. 2015.
\newblock \href {http://arxiv.org/abs/1503.02531} {Distilling the knowledge in
  a neural network}.

\bibitem[{Horne et~al.(2018)Horne, Dron, Khedr, and Adali}]{horne2018sampling}
Benjamin~D. Horne, William Dron, Sara Khedr, and Sibel Adali. 2018.
\newblock Sampling the news producers: A large news and feature data set for
  the study of the complex media landscape.
\newblock In \emph{Proceedings of the International AAAI Conference on Web and
  Social Media (ICWSM)}, Stanford, CA.

\bibitem[{Jin et~al.(2022)Jin, Lalwani, Vaidhya, Shen, Ding, Lyu, Sachan,
  Mihalcea, and Schoelkopf}]{jin-etal-2022-logical}
Zhijing Jin, Abhinav Lalwani, Tejas Vaidhya, Xiaoyu Shen, Yiwen Ding, Zhiheng
  Lyu, Mrinmaya Sachan, Rada Mihalcea, and Bernhard Schoelkopf. 2022.
\newblock \href {https://aclanthology.org/2022.findings-emnlp.532} {Logical
  fallacy detection}.
\newblock In \emph{Findings of the Association for Computational Linguistics:
  EMNLP 2022}, pages 7180--7198, Abu Dhabi, United Arab Emirates. Association
  for Computational Linguistics.

\bibitem[{Kiesel et~al.(2019)Kiesel, Mestre, Shukla, Vincent, Adineh, Corney,
  Stein, and Potthast}]{kiesel2019semeval}
Johannes Kiesel, Maria Mestre, Rishabh Shukla, Emmanuel Vincent, Payam Adineh,
  David Corney, Benno Stein, and Martin Potthast. 2019.
\newblock Semeval-2019 task 4: Hyperpartisan news detection.
\newblock In \emph{Proceedings of the 13th International Workshop on Semantic
  Evaluation}, pages 829--839.

\bibitem[{Lasswell(1927)}]{lasswell1927theory}
Harold~D Lasswell. 1927.
\newblock The theory of political propaganda.
\newblock \emph{American Political Science Review}, 21(3):627--631.

\bibitem[{Lei and Huang(2022)}]{lei-huang-2022-shot}
Yuanyuan Lei and Ruihong Huang. 2022.
\newblock \href {https://doi.org/10.18653/v1/2022.findings-emnlp.409} {Few-shot
  (dis)agreement identification in online discussions with regularized and
  augmented meta-learning}.
\newblock In \emph{Findings of the Association for Computational Linguistics:
  EMNLP 2022}, pages 5581--5593, Abu Dhabi, United Arab Emirates. Association
  for Computational Linguistics.

\bibitem[{Lei et~al.(2022)Lei, Huang, Wang, and
  Beauchamp}]{lei-etal-2022-sentence}
Yuanyuan Lei, Ruihong Huang, Lu~Wang, and Nick Beauchamp. 2022.
\newblock \href {https://doi.org/10.18653/v1/2022.emnlp-main.682}
  {Sentence-level media bias analysis informed by discourse structures}.
\newblock In \emph{Proceedings of the 2022 Conference on Empirical Methods in
  Natural Language Processing}, pages 10040--10050, Abu Dhabi, United Arab
  Emirates. Association for Computational Linguistics.

\bibitem[{Li et~al.(2019)Li, Zhang, Si, and Liu}]{li-etal-2019-rumor}
Quanzhi Li, Qiong Zhang, Luo Si, and Yingchi Liu. 2019.
\newblock \href {https://doi.org/10.18653/v1/D19-5008} {Rumor detection on
  social media: Datasets, methods and opportunities}.
\newblock In \emph{Proceedings of the Second Workshop on Natural Language
  Processing for Internet Freedom: Censorship, Disinformation, and Propaganda},
  pages 66--75, Hong Kong, China. Association for Computational Linguistics.

\bibitem[{Little(2017)}]{little2017propaganda}
Andrew~T Little. 2017.
\newblock Propaganda and credulity.
\newblock \emph{Games and Economic Behavior}, 102:224--232.

\bibitem[{Liu et~al.(2022)Liu, Zhang, Wegsman, Beauchamp, and
  Wang}]{liu-etal-2022-politics}
Yujian Liu, Xinliang~Frederick Zhang, David Wegsman, Nicholas Beauchamp, and
  Lu~Wang. 2022.
\newblock \href {https://doi.org/10.18653/v1/2022.findings-naacl.101}
  {{POLITICS}: Pretraining with same-story article comparison for ideology
  prediction and stance detection}.
\newblock In \emph{Findings of the Association for Computational Linguistics:
  NAACL 2022}, pages 1354--1374, Seattle, United States. Association for
  Computational Linguistics.

\bibitem[{Loshchilov and Hutter(2019)}]{loshchilov2019decoupled}
Ilya Loshchilov and Frank Hutter. 2019.
\newblock \href {http://arxiv.org/abs/1711.05101} {Decoupled weight decay
  regularization}.

\bibitem[{Miller and Robinson(2019)}]{miller2019propaganda}
David Miller and Piers Robinson. 2019.
\newblock Propaganda, politics and deception.
\newblock \emph{The Palgrave handbook of deceptive communication}, pages
  969--988.

\bibitem[{Oshikawa et~al.(2020)Oshikawa, Qian, and
  Wang}]{oshikawa-etal-2020-survey}
Ray Oshikawa, Jing Qian, and William~Yang Wang. 2020.
\newblock \href {https://aclanthology.org/2020.lrec-1.747} {A survey on natural
  language processing for fake news detection}.
\newblock In \emph{Proceedings of the Twelfth Language Resources and Evaluation
  Conference}, pages 6086--6093, Marseille, France. European Language Resources
  Association.

\bibitem[{Park et~al.(2019)Park, Kim, Lu, and Cho}]{park2019relational}
Wonpyo Park, Dongju Kim, Yan Lu, and Minsu Cho. 2019.
\newblock \href {http://arxiv.org/abs/1904.05068} {Relational knowledge
  distillation}.

\bibitem[{P{\'e}rez-Rosas et~al.(2018)P{\'e}rez-Rosas, Kleinberg, Lefevre, and
  Mihalcea}]{perez-rosas-etal-2018-automatic}
Ver{\'o}nica P{\'e}rez-Rosas, Bennett Kleinberg, Alexandra Lefevre, and Rada
  Mihalcea. 2018.
\newblock \href {https://aclanthology.org/C18-1287} {Automatic detection of
  fake news}.
\newblock In \emph{Proceedings of the 27th International Conference on
  Computational Linguistics}, pages 3391--3401, Santa Fe, New Mexico, USA.
  Association for Computational Linguistics.

\bibitem[{Peters et~al.(2018)Peters, Neumann, Iyyer, Gardner, Clark, Lee, and
  Zettlemoyer}]{peters-etal-2018-deep}
Matthew~E. Peters, Mark Neumann, Mohit Iyyer, Matt Gardner, Christopher Clark,
  Kenton Lee, and Luke Zettlemoyer. 2018.
\newblock \href {https://doi.org/10.18653/v1/N18-1202} {Deep contextualized
  word representations}.
\newblock In \emph{Proceedings of the 2018 Conference of the North {A}merican
  Chapter of the Association for Computational Linguistics: Human Language
  Technologies, Volume 1 (Long Papers)}, pages 2227--2237, New Orleans,
  Louisiana. Association for Computational Linguistics.

\bibitem[{Prasad et~al.(2008)Prasad, Dinesh, Lee, Miltsakaki, Robaldo, Joshi,
  and Webber}]{prasad-etal-2008-penn}
Rashmi Prasad, Nikhil Dinesh, Alan Lee, Eleni Miltsakaki, Livio Robaldo,
  Aravind Joshi, and Bonnie Webber. 2008.
\newblock \href
  {http://www.lrec-conf.org/proceedings/lrec2008/pdf/754_paper.pdf} {The {P}enn
  {D}iscourse {T}ree{B}ank 2.0.}
\newblock In \emph{Proceedings of the Sixth International Conference on
  Language Resources and Evaluation ({LREC}'08)}, Marrakech, Morocco. European
  Language Resources Association (ELRA).

\bibitem[{Rashkin et~al.(2017)Rashkin, Choi, Jang, Volkova, and
  Choi}]{rashkin2017truth}
Hannah Rashkin, Eunsol Choi, Jin~Yea Jang, Svitlana Volkova, and Yejin Choi.
  2017.
\newblock Truth of varying shades: Analyzing language in fake news and
  political fact-checking.
\newblock In \emph{Proceedings of the 2017 Conference on Empirical Methods in
  Natural Language Processing}, pages 2931--2937, Copenhagen, Denmark.
  Association for Computational Linguistics.

\bibitem[{Rubin et~al.(2016)Rubin, Conroy, Chen, and
  Cornwell}]{rubin-etal-2016-fake}
Victoria Rubin, Niall Conroy, Yimin Chen, and Sarah Cornwell. 2016.
\newblock \href {https://doi.org/10.18653/v1/W16-0802} {Fake news or truth?
  using satirical cues to detect potentially misleading news}.
\newblock In \emph{Proceedings of the Second Workshop on Computational
  Approaches to Deception Detection}, pages 7--17, San Diego, California.
  Association for Computational Linguistics.

\bibitem[{Stanley(2015)}]{stanley2015propaganda}
Jason Stanley. 2015.
\newblock How propaganda works.
\newblock In \emph{How propaganda works}. Princeton University Press.

\bibitem[{Tardaguila et~al.(2018)Tardaguila, Benevenuto, and
  Ortellado}]{tardaguila2018fake}
Cristina Tardaguila, Fabricio Benevenuto, and Pablo Ortellado. 2018.
\newblock \href
  {https://www.nytimes.com/2018/10/17/opinion/brazil-election-fake-news-whatsapp.html}
  {Fake news is poisoning brazilian politics. whatsapp can stop it}.
\newblock \emph{The New York Times}.

\bibitem[{van~den Berg and Markert(2020)}]{van2020context}
Esther van~den Berg and Katja Markert. 2020.
\newblock Context in informational bias detection.
\newblock In \emph{Proceedings of the 28th International Conference on
  Computational Linguistics}, pages 6315--6326.

\bibitem[{Vaswani et~al.(2017)Vaswani, Shazeer, Parmar, Uszkoreit, Jones,
  Gomez, Kaiser, and Polosukhin}]{vaswani2017attention}
Ashish Vaswani, Noam Shazeer, Niki Parmar, Jakob Uszkoreit, Llion Jones,
  Aidan~N. Gomez, Lukasz Kaiser, and Illia Polosukhin. 2017.
\newblock \href {http://arxiv.org/abs/1706.03762} {Attention is all you need}.

\bibitem[{Vijayaraghavan and
  Vosoughi(2022)}]{vijayaraghavan-vosoughi-2022-tweetspin}
Prashanth Vijayaraghavan and Soroush Vosoughi. 2022.
\newblock \href {https://doi.org/10.18653/v1/2022.naacl-main.251} {{TWEETSPIN}:
  Fine-grained propaganda detection in social media using multi-view
  representations}.
\newblock In \emph{Proceedings of the 2022 Conference of the North American
  Chapter of the Association for Computational Linguistics: Human Language
  Technologies}, pages 3433--3448, Seattle, United States. Association for
  Computational Linguistics.

\bibitem[{Vlad et~al.(2019)Vlad, Tanase, Onose, and
  Cercel}]{vlad-etal-2019-sentence}
George-Alexandru Vlad, Mircea-Adrian Tanase, Cristian Onose, and
  Dumitru-Clementin Cercel. 2019.
\newblock \href {https://doi.org/10.18653/v1/D19-5022} {Sentence-level
  propaganda detection in news articles with transfer learning and
  {BERT}-{B}i{LSTM}-capsule model}.
\newblock In \emph{Proceedings of the Second Workshop on Natural Language
  Processing for Internet Freedom: Censorship, Disinformation, and Propaganda},
  pages 148--154, Hong Kong, China. Association for Computational Linguistics.

\bibitem[{Wei et~al.(2021)Wei, Hu, Zhou, Yue, and Hu}]{wei-etal-2021-towards}
Lingwei Wei, Dou Hu, Wei Zhou, Zhaojuan Yue, and Songlin Hu. 2021.
\newblock \href {https://doi.org/10.18653/v1/2021.acl-long.297} {Towards
  propagation uncertainty: Edge-enhanced {B}ayesian graph convolutional
  networks for rumor detection}.
\newblock In \emph{Proceedings of the 59th Annual Meeting of the Association
  for Computational Linguistics and the 11th International Joint Conference on
  Natural Language Processing (Volume 1: Long Papers)}, pages 3845--3854,
  Online. Association for Computational Linguistics.

\bibitem[{Williams(1992)}]{williams1992simple}
Ronald~J Williams. 1992.
\newblock Simple statistical gradient-following algorithms for connectionist
  reinforcement learning.
\newblock \emph{Machine learning}, 8:229--256.

\bibitem[{Yu et~al.(2021)Yu, Da~San~Martino, Mohtarami, Glass, and
  Nakov}]{yu-etal-2021-interpretable}
Seunghak Yu, Giovanni Da~San~Martino, Mitra Mohtarami, James Glass, and Preslav
  Nakov. 2021.
\newblock \href {https://aclanthology.org/2021.ranlp-1.179} {Interpretable
  propaganda detection in news articles}.
\newblock In \emph{Proceedings of the International Conference on Recent
  Advances in Natural Language Processing (RANLP 2021)}, pages 1597--1605, Held
  Online. INCOMA Ltd.

\end{thebibliography}
\bibliographystyle{acl_natbib}

\appendix

\newpage

\section{Appendix}

\subsection{ChatGPT Prompt}
\label{prompt}

The designed instruction prompt for sentence-level propaganda identification task is: "Propaganda is a form of misinformation or deceptive narratives that incite or mislead the public, usually with a political purpose. Please reply Yes if the following sentence contains propaganda content, else reply No. Sentence: "xxx". Answer:"

The designed instruction prompt for token-level propaganda identification task is: "Propaganda is a form of misinformation or deceptive narratives that incite or mislead the public, usually with a political purpose. Please extract the word in the following sentences that contains propaganda content. Please mimic the following output style. Example: "Of course, no "mistake" had occurred, the ban has been lifted only because of the wide publicity that we engaged in.". Words: wide, publicity. Sentence: "xxx". Words:"

\end{document}